\documentclass[10pt,twocolumn,letterpaper]{article}

\usepackage{cvpr}              
\usepackage{graphicx}

\definecolor{cvprblue}{rgb}{0.21,0.49,0.74}
\usepackage[pagebackref,breaklinks,colorlinks,allcolors=cvprblue]{hyperref}

\def\paperID{00004} 
\def\confName{CVPR}
\def\confYear{2025}

\title{AI Hiring with LLMs: A Context-Aware and Explainable Multi-Agent Framework for Resume Screening}

\author{
Frank P.-W. Lo$^{1}$ \quad
Jianing Qiu$^{2}$ \quad
Zeyu Wang$^{1}$ \quad
Haibao Yu$^{3}$\\
Yeming Chen$^{4}$ \quad
Gao Zhang$^{5}$ \quad
Benny Lo$^{1}$\\[0.5em]
$^{1}$Imperial College London\quad
$^{2}$The Chinese University of Hong Kong\\
$^{3}$The University of Hong Kong\quad
$^{4}$Wedon Education Technologies\quad
$^{5}$Brest Business School\\[0.5em]
\footnotesize\texttt{\{po.lo15, zeyu.wang20, benny.lo\}@imperial.ac.uk, jianingqiu@cuhk.edu.hk,}\\
\footnotesize\texttt{yuhaibao94@gmail.com, chenym@wedon.com, gao.zhang@brest-bs.com}
}

\begin{document}
\maketitle
\begin{abstract}
\textcolor{black}{Resume screening is a critical yet time-intensive process in talent acquisition, requiring recruiters to analyze vast volume of job applications while remaining objective, accurate, and fair. With the advancements in Large Language Models (LLMs), their reasoning capabilities and extensive knowledge bases demonstrate new opportunities to streamline and automate recruitment workflows. In this work, we propose a multi-agent framework for resume screening using LLMs to systematically process and evaluate resumes. The framework consists of four core agents, including a resume extractor, an evaluator, a summarizer, and a score formatter. To enhance the contextual relevance of candidate assessments, we integrate Retrieval-Augmented Generation (RAG) within the resume evaluator, allowing incorporation of external knowledge sources, such as industry-specific expertise, professional certifications, university rankings, and company-specific hiring criteria. This dynamic adaptation enables personalized recruitment, bridging the gap between AI automation and talent acquisition. We assess the effectiveness of our approach by comparing AI-generated scores with ratings provided by HR professionals on a dataset of anonymized online resumes. The findings highlight the potential of multi-agent RAG-LLM systems in automating resume screening, enabling more efficient and scalable hiring workflows.} 


\end{abstract}

\section{Introduction}

\begin{figure}[t]

    \centering
    \includegraphics[width=\linewidth]{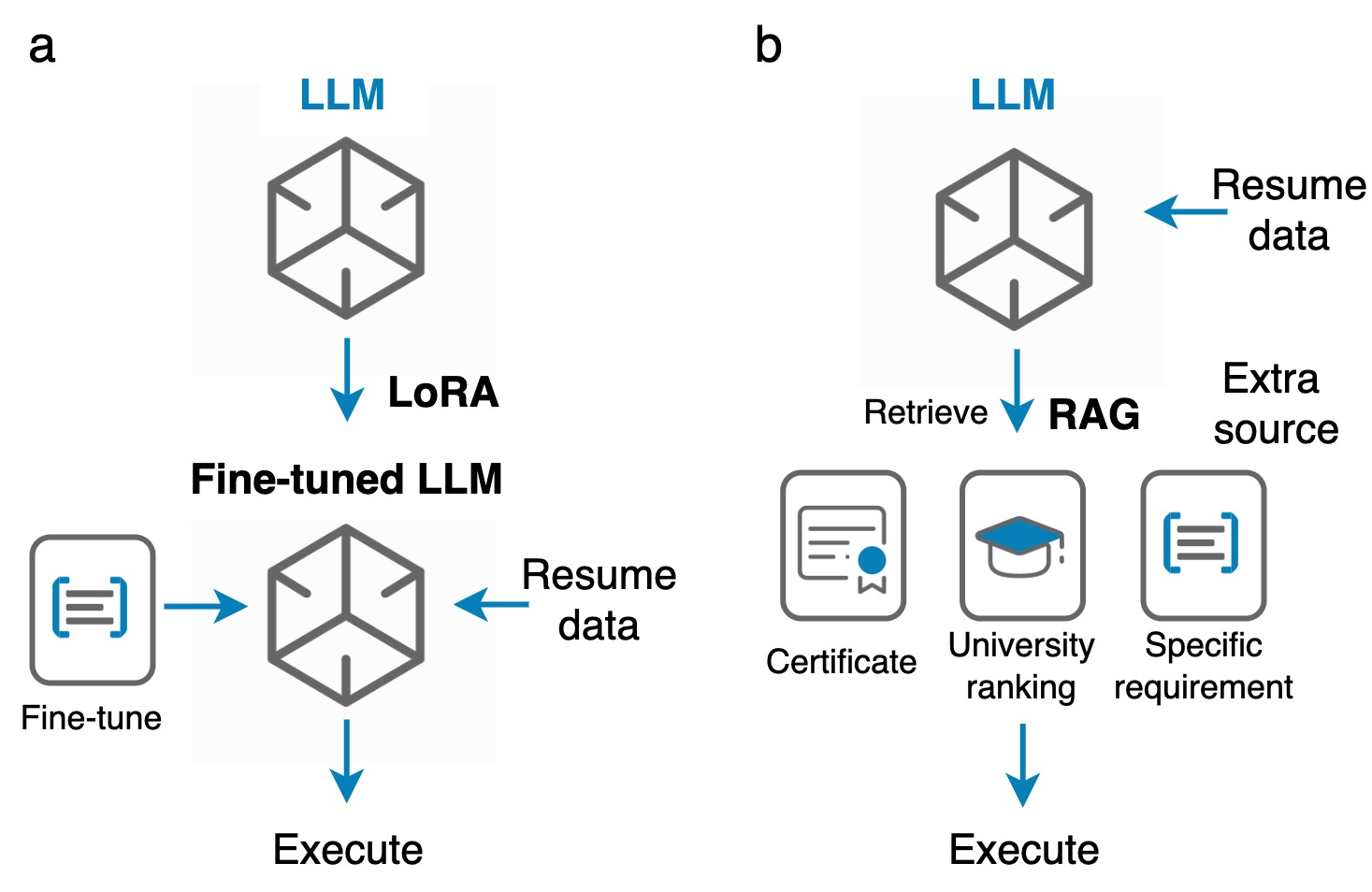}
    \caption{\textbf{Illustration diagram of fine-tuned LLM and RAG-LLM for resume screening.} (a) Traditional fine-tuning approaches (e.g., LoRA) require updating model parameters to adapt to new tasks (i.e., new companies' hiring requirements). (b) Our model uses RAG, eliminating the need for fine-tuning by dynamically retrieving relevant information from external sources. }
    \label{lora}
\vspace{-10pt}
\end{figure}

Automated resume screening is a critical component of the hiring process. Companies often receive a high volume of job applications, making it difficult to manually review every resume or CV efficiently. Traditional resume screening methods primarily rely on rule-based approaches, and keyword matching, which often fail to include job-specific requirements and lack adaptability. In addition, such methods provide limited transparency and feedback, making it difficult for recruiters to interpret and validate AI-driven decisions. Traditional methods also face several technical challenges, including difficulties in comprehending the nuanced choice of words in a resume and accurately interpreting the syntax of unstructured written language \cite{sinha2021resume}. Most importantly, resume screening systems are expected to keep up with the constantly changing job market and business needs. Therefore, models need to be updated frequently as new job opportunities emerge. For instance, a recommendation that was relevant last month might become obsolete if the job market shifts, such as due to a sudden surge in demand for specific skills. Hence, the integration of real-time data and continuous learning is still a key focus of ongoing research.

\begin{figure*}[t]
    \centering
    \includegraphics[width=0.95\linewidth]{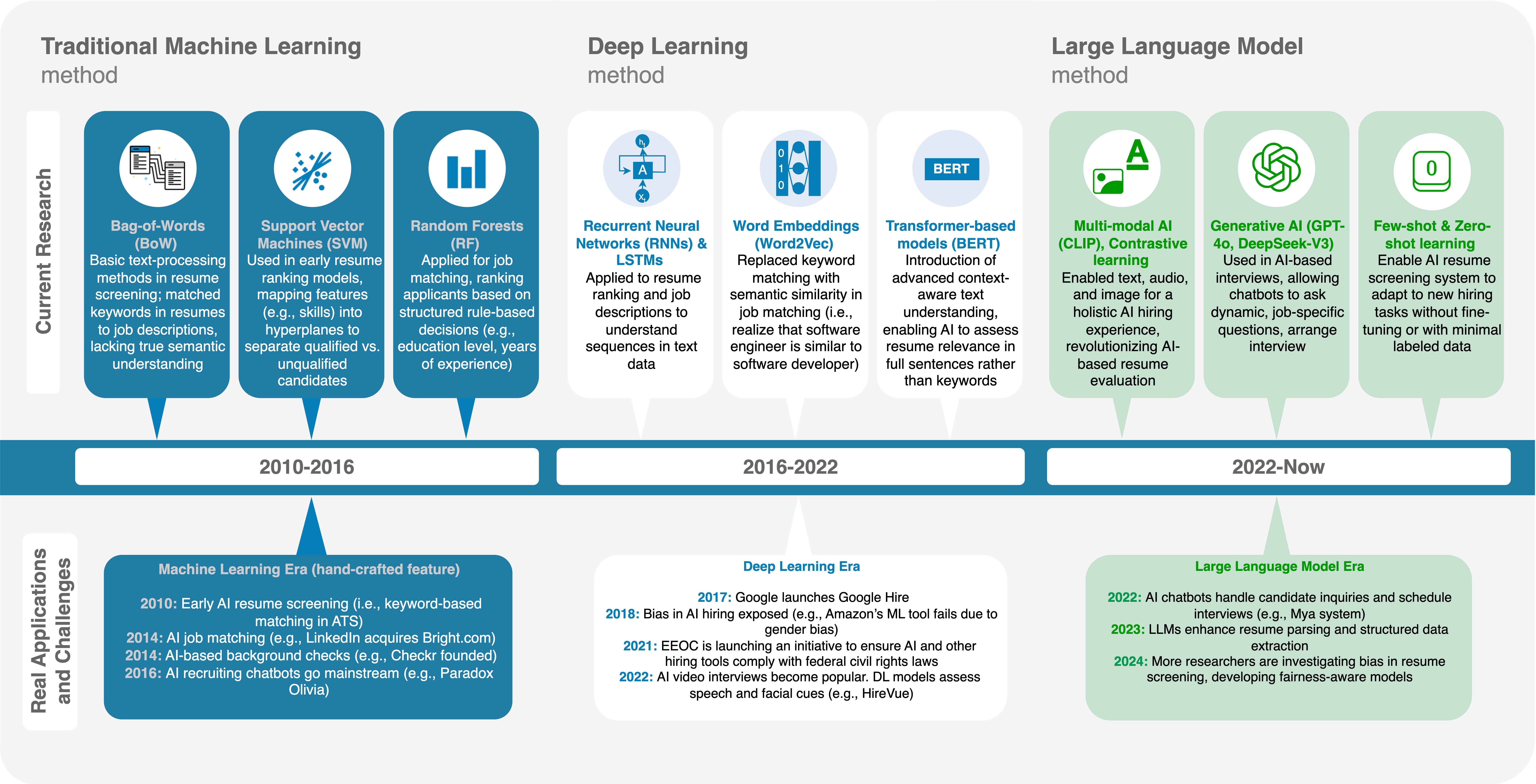}
    \caption{\textbf{The evolution of AI-driven hiring technologies.} This figure presents the transition of AI-driven hiring methods across three major eras: traditional machine learning (2010-2016), deep learning (2016-2022), and large language models (2022-present). It highlights key advancements in AI hiring technologies and notable case studies demonstrating their real-world applications \cite{LinkedInBrightAcquisition, ForbesCheckr, SAP2019, GoogleHire2017, samadhiya2022importance}. }
    \label{evoluation}
    \vspace{-10pt}
\end{figure*}

Recent advancements in LLMs have demonstrated remarkable reasoning capabilities \cite{sun2023survey}, enabling new possibilities for intelligent resume screening. However, most existing LLM-driven screening systems operate as monolithic models \cite{bouleanu2024benefits}, where resume parsing, evaluation, and feedback generation are handled in a single-step process (i.e., single LLM). The major drawback of single LLM approaches is their lack of modularity. Since resume extraction, evaluation, and feedback generation are coupled in a single model call, modifying the scoring logic requires retraining or fine-tuning the entire model (e.g., using Low-Rank Adaptation (LoRA)) as shown in Figure \ref{lora}(a). This makes it difficult to adapt screening criteria across different industries and job roles, reducing overall scalability\footnote{In typical recruitment workflows, candidates are evaluated by HR across multiple dimensions, such as experience, skills, and education, based on job-specific scoring criteria, which can vary significantly across companies and roles.}.  Additionally, when multiple reasoning steps (e.g., extracting information from resume, applying scoring criteria, and justifying decisions) are handled simultaneously, it becomes difficult to interpret the decision-making process. This lack of transparency limits recruiters' ability to validate AI-driven evaluations and adjust the system without extensive reconfiguration.

To address these challenges, we propose a multi-agent framework for resume screening, leveraging Retrieval-Augmented Generation enhanced LLMs (RAG-LLMs) \cite{lewis2020retrieval} within an agentic architecture \cite{qiu2024llm}. Unlike single-step models, our framework consists of four core agents, each responsible for a distinct function: resume extraction (hiring assistant agent), evaluation (hiring manager agent), summarization (hiring coordinator agent), and score formatting (data curator agent). This modular structure provides greater flexibility, and in our design, the evaluation agent can dynamically retrieve company-specific hiring criteria via RAG. Instead of requiring fine-tuning, the system can adjust its evaluation standards in real-time by allowing HR professionals to upload job requirement documents to the backend, making it highly adaptable across different industries and job roles as shown in Figure \ref{lora}(b). Moreover, by dividing the screening process into multiple independent agents, the framework enhances transparency and explainability. Each stage of the process remains clearly defined, allowing recruiters to trace how a candidate was assessed and why a particular score was assigned (i.e., instead of outputting a score alone, the evaluation criteria can be inferred from the extracted resume content and the generated feedback, resulting in more meaningful and explainable outcomes). This also ensures that changes in scoring criteria does not interfere with data extraction and feedback generation. By leveraging multi-agent modularity and RAG-based dynamic retrieval, our framework provides a scalable, transparent, and adaptable solution for AI-driven resume screening. Figure \ref{evoluation} illustrates how AI-driven hiring technologies have evolved to address other challenges as well. The contributions of this paper are summarized as follows:


\begin{itemize}
    \item We propose a multi-agent architecture that introduces a modular structure, enhancing explainability and transparency in resume screening.
    
    \item Our framework is designed to adapt to diverse hiring criteria across different roles (e.g., leadership skills for department directors, HR expertise for human resource associates), enabling a more context-aware and role-adaptive screening process.
    
    \item By integrating RAG, our system allows recruiters to dynamically adjust screening parameters (e.g., prioritizing specific university rankings, certifications, or domain expertise) without requiring LLM retraining/fine-tuning, thereby enhancing adaptability and customization.
    

    \item We discuss the future of AI in hiring, addressing ethical considerations, bias mitigation, and regulatory challenges, while also examining how LLMs can enhance fairness and efficiency in recruitment.
\end{itemize}

\section{Related Work}

\subsection{AI-driven hiring}

The adoption of AI in hiring has significantly transformed recruitment processes, enabling automation in resume screening \cite{gan2024application}, resume classification \cite{pal2022resume, ali2022resume, jiechieu2021skills, jalili2024bilstm, nasser2018convolutional, ramraj2020real, skondras2023generating, bharadwaj2022resume}, resume ranking \cite{satheesh2020resume, tejaswini2022design}, interview evaluation \cite{mirowska2022preferring, uppalapati2025ai, kim2023fairness, rk2025ai}, salary prediction \cite{das2020salary, dutta2018design}, and also bias mitigation \cite{deshpande2020mitigating, wilson2024gender, chen2018investigating, mujtaba2019ethical}. With the emergence of ML, DL, and LLMs, AI-driven hiring systems have evolved from simple keyword-based matching to context-aware decision-making.

\subsection{Resume screening systems}

Early AI-driven resume screening systems primarily relied on traditional machine learning methods, such as Bag-of-Words (BoW), Support Vector Machines (SVM), and Random Forests (RF) \cite{pal2022resume}. These methods treated resumes as structured data, applying rule-based decision-making to assess candidate qualifications. However, these models lacked semantic understanding and relied solely on keyword matching (e.g., failing to recognize that software developer is equivalent to software engineer), leading to high error rates in candidate selection. The transition from traditional machine learning to deep learning marked a significant shift in resume screening, enabling models to process sequential and semantic text information. Convolutional Neural Networks (CNNs), Recurrent Neural Networks (RNNs) and Long Short-Term Memory (LSTMs) \cite{jalili2024bilstm} were among the first deep learning models applied to resume screening, improving accuracy by capturing sequential dependencies in text data. Further advancements introduced word embeddings (e.g., Word2Vec \cite{church2017word2vec}), which replaced keyword matching with semantic similarity, allowing AI to recognize that terms like software engineer and software developer are contextually related. However, these embeddings are context-independent. More recently, transformer-based models (e.g., BERT \cite{devlin2019bert}) introduced context-aware text understanding, enabling AI to assess resume relevance in full-sentence representations rather than isolated keywords. While deep learning improved resume parsing, job matching, and ranking, these models required large-scale labeled data for training, limiting their adaptability across diverse hiring contexts.
The advancements of LLMs have transformed AI-driven resume screening, enabling zero-shot and few-shot learning to assess candidates without extensive labeled training data. Unlike traditional machine learning and early deep learning models, LLMs could leverage large-scale pre-training to extract key resume attributes, analyze job relevance, and infer contextual qualifications dynamically. Prior advancements, such as Word2Vec and BERT, already improved semantic resume-job matching, reducing reliance on exact keyword matches. However, LLMs further enhance contextual reasoning, allowing for deeper candidate evaluation, such as identifying transferable skills (i.e., recognizing transferable skills such as proficiency in C++ from experience with embedded systems or Python from data analysis projects, even if not explicitly stated in the resume) and inferring implicit qualifications. Nevertheless, there are limited studies on applying LLMs to resume screening \cite{gan2024application, salakar2023resume, haryan2024fairhire}, and key challenges remain. For instances, LLMs rely on static pretraining data, restricting their ability to adapt to dynamic hiring criteria and evolving job requirements as mentioned.

\subsection{LLMs with RAG}
RAG has been widely explored in fields like customer support \cite{xu2024retrieval}, legal research \cite{wiratunga2024cbr}, medicine \cite{zakka2024almanac, xiong2024benchmarking}, finance \cite{yepes2024financial}, and education \cite{levonian2023retrieval, wei2024medco}, enhancing LLMs by integrating real-time, domain-specific information retrieval. It improves accuracy, reduces hallucinations, and enables context-aware decision-making \cite{li2022survey}. However, its application in resume screening remains limited, with most AI resume screening systems relying on static embeddings or rule-based models. Exploring RAG-enhanced resume screening could improve hiring procedure by integrating real-time labor market data and hiring trends, offering a more adaptive and intelligent screening process.

\begin{figure*}[t]
    \centering
    \includegraphics[width=\linewidth]{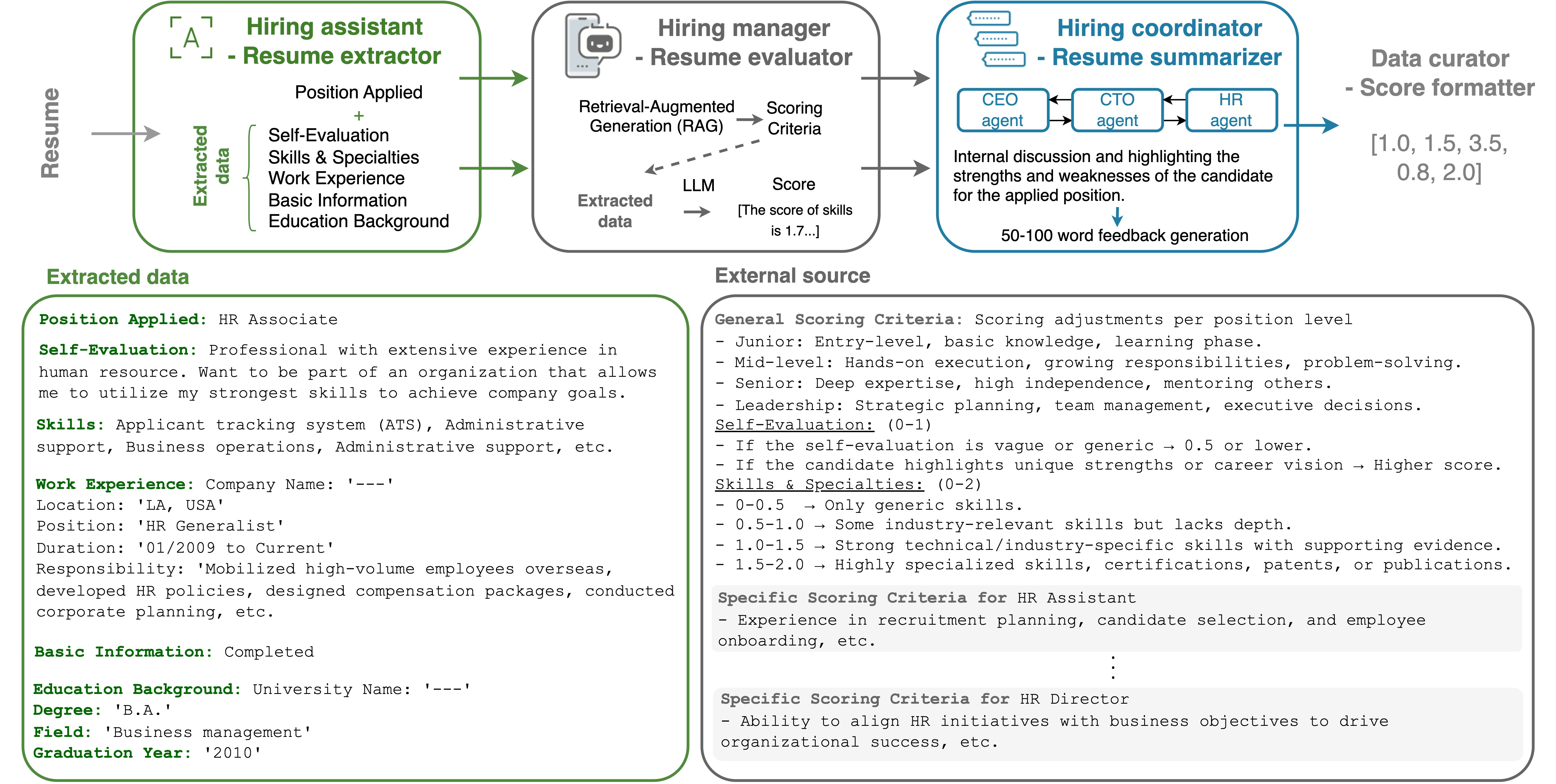}
    \caption{\textbf{Illustration of the proposed multi-agent framework for resume screening.} The framework consists of four core agents: Resume extractor, responsible for parsing and structuring resume content; Resume evaluator, which assigns scores based on predefined criteria while integrating external knowledge via RAG; Resume summarizer, which consists of three sub-agents that generate feedback through collective decision-making, ensuring a comprehensive evaluation of the candidate's strengths and weaknesses; Score formatter, which organizes evaluation results into a structured format for future analysis. This modular approach enhances explainability and adaptability, as recruiters can review each step of the evaluation process without requiring to examine the raw resume directly.}
    \label{agent}
    \vspace{-10pt}
\end{figure*}



\section{Problem Definition}

Our framework enables dynamic, context-aware resume screening by adapting evaluation scores based on the applied job role. Unlike traditional models with fixed evaluation scores, our framework assesses candidates using role-specific standards. Given a resume \( R \), the model generates a score vector:


\begin{equation}
S^J = \{ S_S^J, S_K^J, S_W^J, S_B^J, S_E^J \}
\end{equation}
where $S_S^J$ refers to the score of \text{self-evaluation}, $S_K^J$ refers to the score of \text{skills \& specialties}, $S_W^J$ refers to the score of \text{work experience}, $S_B^J$ refers to the score of \text{basic information}, $S_E^J$ refers to the score of \text{education background} and $J$ refers to the applied job position. Each evaluation criterion is assigned a fixed weight across all job roles:

\begin{equation}
W = \{ w_S, w_K, w_W, w_B, w_E \}, \quad \sum w_i = 1
\end{equation}
where \( w_i \) remains constant regardless of the applied job position. Besides, one of the most challenging aspects of this work is role-specific scoring. A candidate applying for a HR intern versus a HR director should receive different scores, even with the same resume:

\begin{equation}
S_W^{\text{Intern}} > S_W^{\text{Director}} \quad \text{(for low-experience candidates)}
\end{equation}
Different aspects of evaluation should be interpreted contextually based on job requirements. In our work, the LLM generates job-specific evaluation scores via job applied $J$:

\begin{equation}
S^J = \text{LLM}(R, J).
\end{equation}
The final score for a candidate is computed as:

\begin{equation}
S_{\text{final}}^J = \sum w_i S_i^J,
\end{equation}

\section{Detailed Information and Methodology}

The proposed framework streamlines resume screening using a multi-agent approach, where different components work together to analyze and evaluate job applications efficiently. The framework consists of four core agents: the resume extractor, resume evaluator, resume summarizer, and score formatter, each handling a specific task in the process as shown in Figure \ref{agent}. First, the resume extractor identifies key details from a candidate’s resume. Since resumes come in different formats, this step ensures that all information is structured in a clear and standardized way. Next, the resume evaluator reviews the extracted details and assigns scores based on how well the candidate’s qualifications match the job requirements. The resume summarizer then generates a concise, easy-to-understand report, highlighting the candidate’s strengths and areas for improvement. This helps recruiters quickly assess candidates without going through lengthy resumes. Finally, the score formatter standardizes the evaluation output into a structured numerical format. This ensures consistency in how candidate scores are presented, making it easier to compare applicants and integrate results into decision-making systems.

\subsection{Resume extractor agent}

The extractor agent, acting as the hiring assistant, leverages reasoning capabilities of LLM to extract structured information from unstructured text accurately, ensuring the precise identification of key details such as the 1) position applied for (i.e., position name and its level: junior, mid-level, senior, or leadership) 2) self-evaluation 3) skills \& specialties 4) work experience (i.e., company name, duration, and responsibilities) 5) basic information, and 6) education background. Unlike traditional keyword-based extraction methods, the LLM processes unstructured text with contextual understanding, allowing it to infer missing details, and recognize implicit skills.

\subsection{Resume evaluator agent}

The evaluator agent, functioning as a hiring manager, assigns scores based on five evaluation categories: self-evaluation (score: 0-1), skills \& specialties (score: 0-2), work experience (score: 0-4), basic information (score: 0-1), and educational background (score: 0-2). Instead of relying solely on predefined rules, the evaluator agent leverages RAG to dynamically retrieve company-specific hiring criteria, job descriptions, and other relevant information from an external source. The details of the RAG pipeline can be structured as follows:

\subsubsection{Vector embedding}

All document (i.e., external source) chunks are encoded into dense vector representations using an embedding function \( f_{\text{embed}} \). Let the original job query (e.g., a job requirement) and document chunks be denoted as $q_\text{text}$ and $d_{i,\text{text}}$.

\begin{equation}
q = f_{\text{embed}}(q_\text{text}), \quad d_i = f_{\text{embed}}(d_{i,\text{text}})
\end{equation}
where $q, d_i \in \mathbb{R}^D$, and $D$ is the embedding dimension (i.e., we use OpenAIEmbeddings to generate dense vector representations and ChromaDB as the vector database).



\subsubsection{Cosine similarity computation}

The relevance of document chunks to the query is quantified using cosine similarity. For the query \( q \) and the \( i \)-th document chunk \( d_i \):

\begin{equation}
\text{sim}(q, d_i) = \frac{q \cdot d_i}{\|q\| \|d_i\|}
\end{equation}
where $\text{sim}(q, d_i)$ is the similarity between the vectors, with higher values indicating greater relevance. A relevance threshold \( \tau = 0.3 \) is used to filter out low-relevance document chunks:

\begin{equation}
\text{sim}(q, d_i) \geq \tau \iff \text{Retrieve } d_i 
\end{equation}
where \( q \) refers to the query and \( d_i \) refers to the document chunks.

\begin{figure}[h]
    \centering
    \includegraphics[width=\linewidth]{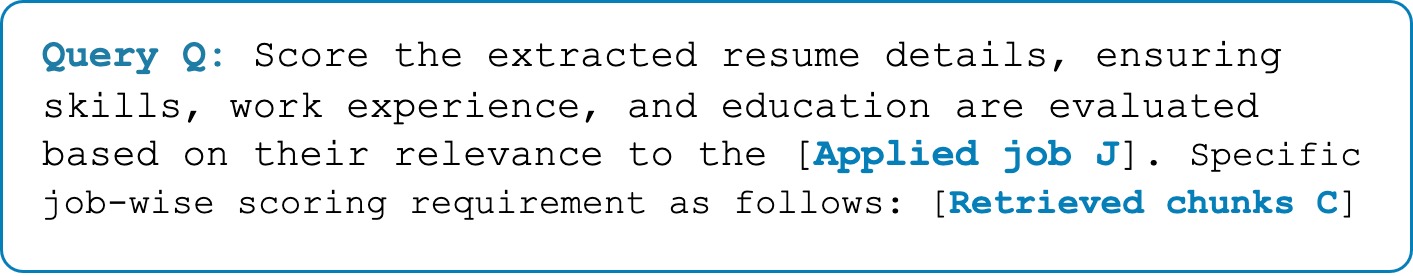}
    \caption{\textbf{Query formulation for resume evaluation agent}. The query instructs the system to score extracted resume details by assessing skills, work experience, and education in relation to the applied job (J). It incorporates retrieved knowledge chunks (C) to ensure job-specific scoring criteria are considered.}
    \label{query}
    \vspace{-5pt}
\end{figure}

\subsubsection{Contextual prompt construction}
Retrieved chunks are formatted into a structured input prompt \( P \) as follows:

\begin{equation}
P = \text{concat}(Q, J, C)  
\end{equation}

\begin{equation}
C = d_1^{\text{(retrieved)}} \cup d_2^{\text{(retrieved)}} \cup \dots \cup d_i^{\text{(retrieved)}}
\end{equation}
where $C$ is the concatenation of the retrieved document chunks, $Q$ represents the query text, and $J$ denotes the applied job position. The formatted prompt $P$ serves as the input to the evaluator agent, which processes the structured information to assess the candidate’s background against predefined job-specific criteria and assigns a resume score accordingly as shown in Figure \ref{query}.

\begin{table*}[]
\centering
\caption{Comparison of single LLMs and the multi-agent RAG-LLMs with different LLM backbones}
\label{table1}
\resizebox{0.65\textwidth}{!}{%
\begin{tabular}{llccccccc}
 & \multicolumn{1}{c}{Model} & PC$_{20}$$\uparrow$* & SC$_{20}$$\uparrow$ & PC$_{15}$$\uparrow$ & SC$_{15}$$\uparrow$ & PC$_{10}$$\uparrow$ & SC$_{10}$$\uparrow$ & MAE$\downarrow$ \\ \hline
\multicolumn{1}{c|}{\multirow{2}{*}{\textcolor{black}{Single LLM}}} & \multicolumn{1}{c}{GPT-4o} & 0.67 & 0.59 & 0.69 & 0.62 & 0.74 & 0.65 & 1.26 \\
\multicolumn{1}{c|}{} & \multicolumn{1}{c}{DeepSeek-V3} & 0.67 & 0.60 & 0.67 & 0.62 & 0.70 & 0.71 & 1.08 \\ \hline
\multicolumn{1}{l|}{\multirow{2}{*}{RAG-LLM (ours)}} & \multicolumn{1}{c}{GPT-4o} & 0.69 & 0.66 & 0.72 & \textbf{0.70} & 0.80 & 0.74 & 1.05 \\
\multicolumn{1}{l|}{} & DeepSeek-V3 & \textbf{0.70} & \textbf{0.66} & \textbf{0.75} & 0.69 & \textbf{0.84} & \textbf{0.74} & \textbf{0.90} \\ \hline
\end{tabular}%
}
\vspace{7pt}

\begin{minipage}{\textwidth}
\footnotesize *$\uparrow$ indicates that higher values are better, while $\downarrow$ indicates that lower values are better. PC refers to Pearson Correlation, and SC refers to Spearman Correlation. MAE refers to Mean Absolute Error. The number following PC/SC represents the percentage of scores used in the evaluation. For example, PC$_{10}$ evaluates model performance only on the subset of candidates whose ground truth scores lie in the top and bottom 10\% percentiles.
\end{minipage}
\end{table*}

\begin{table}[]
\centering
\caption{Ablation study of multi-agent RAG-LLMs with and without resume extraction agent}
\label{ablation}
\resizebox{\columnwidth}{!}{%
\begin{tabular}{llcccccc}
 & \multicolumn{1}{c}{Model}  & PC$_{20}$ & SC$_{20}$ & PC$_{15}$ & SC$_{15}$ & PC$_{10}$ & SC$_{10}$ \\ \hline
\multicolumn{1}{c|}{\multirow{2}{*}{\begin{tabular}[c]{@{}c@{}}RAG-LLM\\ w/o extract.\end{tabular}}} & \multicolumn{1}{c}{GPT-4o} & 0.63 & 0.66 & 0.70 & \textbf{0.71} & 0.81 & 0.74 \\
\multicolumn{1}{c|}{} & \multicolumn{1}{c}{DS-V3} & 0.65 & 0.63 & 0.69 & 0.68 & 0.80 & \textbf{0.79} \\ \hline
\multicolumn{1}{c|}{\multirow{2}{*}{\begin{tabular}[c]{@{}c@{}}RAG-LLM\\ w/ extract.\end{tabular}}} & \multicolumn{1}{c}{GPT-4o} & 0.69 & 0.66 & 0.72 & 0.70 & 0.80 & 0.74 \\
\multicolumn{1}{l|}{} & \multicolumn{1}{c}{DS-V3} & \textbf{0.70} & \textbf{0.66} & \textbf{0.75} & 0.69 & \textbf{0.84} & 0.74  \\ \hline
\end{tabular}%
}
\begin{minipage}{\columnwidth}
\vspace{2mm}
\footnotesize *DS-V3 refers to DeepSeek-V3
\end{minipage}
\end{table}

\subsubsection{Specific requirements from external sources}
In addition to structured attributes such as university rankings and professional certifications, we further analyze historical resumes of outstanding candidates and incorporate up-to-date skill demands to refine job-specific requirements. By leveraging LLM-driven summarization, we extract key qualifications, skills, and experience patterns from past hires, establishing a dynamic baseline for evaluating different job positions. This approach ensures that screening criteria remain relevant and adaptive to evolving industry needs.

\subsection{Resume summarizer agent}
The summarizer agent functions as an hiring coordinator, generating personalized resume feedback by analyzing a candidate’s profile against job requirements. It consists of three sub-agents including the CEO agent, CTO agent, and HR agent, which engage in an internal discussion to refine the feedback based on the scores provided by the evaluator agent. The CEO agent assesses leadership potential, the CTO agent evaluates technical expertise, and the HR agent focuses on soft skills and cultural fit. Through collaborative reasoning, these sub-agents exchange insights, debate strengths and weaknesses, and produce structured feedback. This multi-agent approach ensures context-aware, balanced, and actionable recommendations, enhancing the adaptability and explainability of AI-driven resume evaluations.

\subsection{Score formatter agent}
\textcolor{black}{The score formatter agent (i.e., acting as the data curator) standardizes the output of candidate evaluations into a structured format (e.g., [1.0, 1.5, 3.5, 0.8, 1.5]), ensuring consistency across different assessment components. It takes raw scores generated by various evaluation agents (e.g., experience, skills, education) and converts them into a uniform numerical array for downstream processing. This structured output enables easy integration with ranking models and decision-making pipelines.}

\section{Experimental Results}

Our LLM-driven resume screening system is implemented using CrewAI, which coordinates multiple AI agents to enable structured and automated resume evaluation. The framework runs on a PC equipped with an NVIDIA A6000 GPU, ensuring efficient processing of large-scale resume data. It integrates LLMs via the OpenRouter API, utilizing models such as DeepSeek-V3, and GPT-4o, with LangChain facilitating seamless interaction between components. For RAG, the system employs ChromaDB as a vector database for efficient semantic search, enabling retrieval of relevant hiring criteria and context-aware job matching. Additionally, OpenAI embeddings are used to generate dense vector representations, enhancing the accuracy of similarity-based retrieval. Note that for users requiring local implementation due to privacy concerns, Ollama can be integrated to facilitate the local execution of LLMs.

\subsection{Dataset}
We evaluated our model on a dataset consisting of 105 fully anonymized online resumes. The dataset was labeled by HR professionals, who assigned scores based on five key aspects: self-evaluation, skills \& specialties, work experience, basic information, and education. To ensure privacy, all personally identifiable information, including names and company names, was removed. The resumes in the dataset correspond to various job positions, primarily in the field of human resources. The job levels can be categorized into four groups: junior, mid-level, senior, and leadership. The junior-level positions include HR intern and HR assistant, while the mid-level roles consist of HR associate and HR specialist. Senior-level positions include HR manager and senior HR, whereas leadership roles encompass HR director and strategic HR partner. 

\subsection{Evaluation metrics}

To evaluate our proposed resume screening system, we employ the following evaluation metrics: a) Pearson correlation measures the linear relationship between the AI-estimated scores and the human reviewer scores. This metric helps evaluate if the AI system assigns scores in a manner similar to human evaluators b) Spearman correlation assesses the rank-based monotonic relationship between AI and human reviewer scores. Unlike Pearson correlation, it captures non-linear relationships by ranking the scores before computing the correlation c) MAE measures the absolute difference between AI predictions and HR scores, capturing the average magnitude of errors. This metric is particularly useful in understanding the extent of AI's deviation from human judgment.

\begin{figure}[htb]
    \centering
    \includegraphics[width=\linewidth]{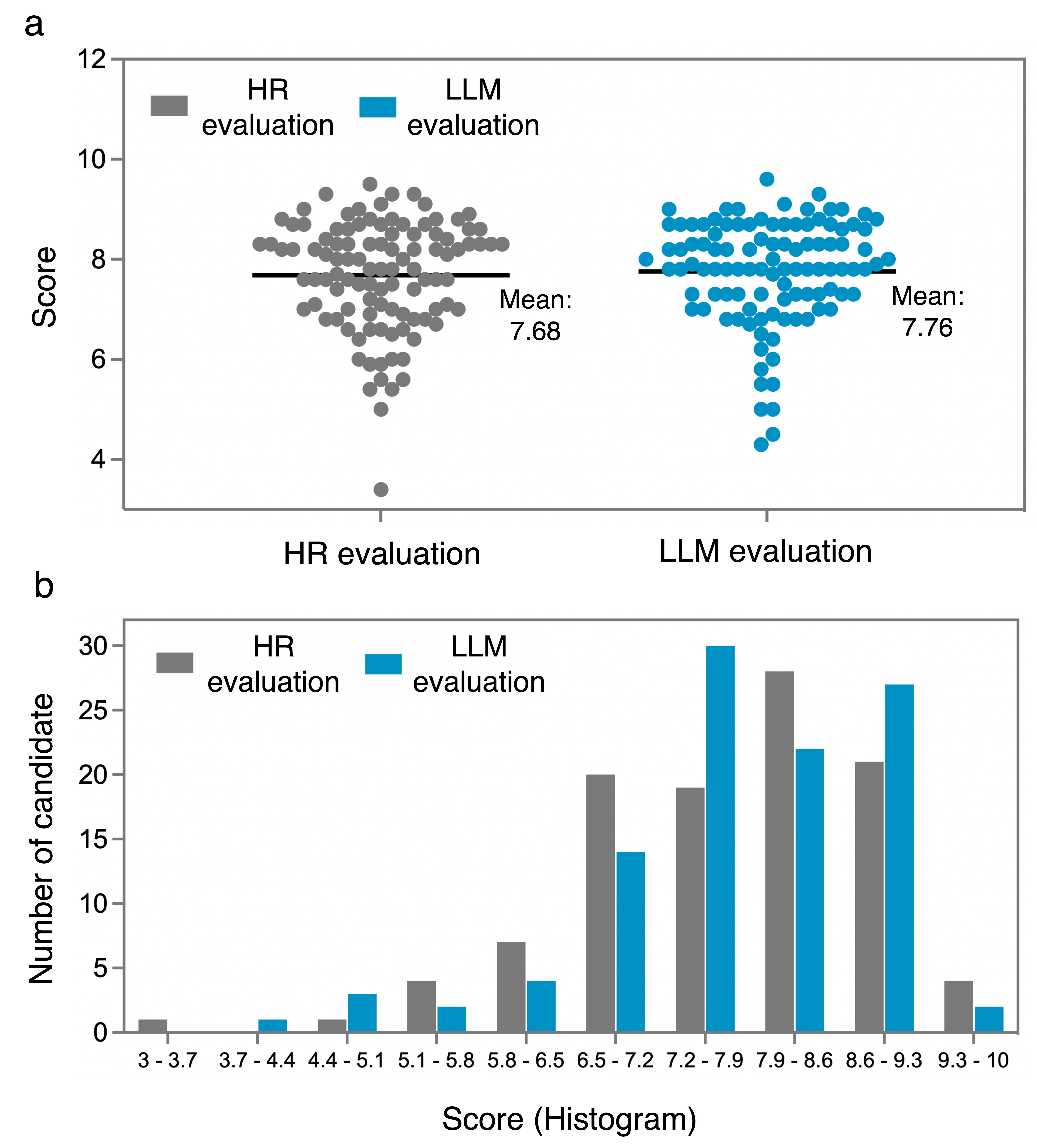}
    \caption{\textbf{Comparison of candidate scores assigned by human evaluators (HR) and a RAG-LLM (DeepSeek-V3).} (a) The scatter plot showing the distribution of scores (b) Histogram showing the number of candidates in each score range based on HR and LLM evaluations.}
    \label{histogram}
\end{figure}

\begin{figure}[htb]
    \centering
    \includegraphics[width=\linewidth]{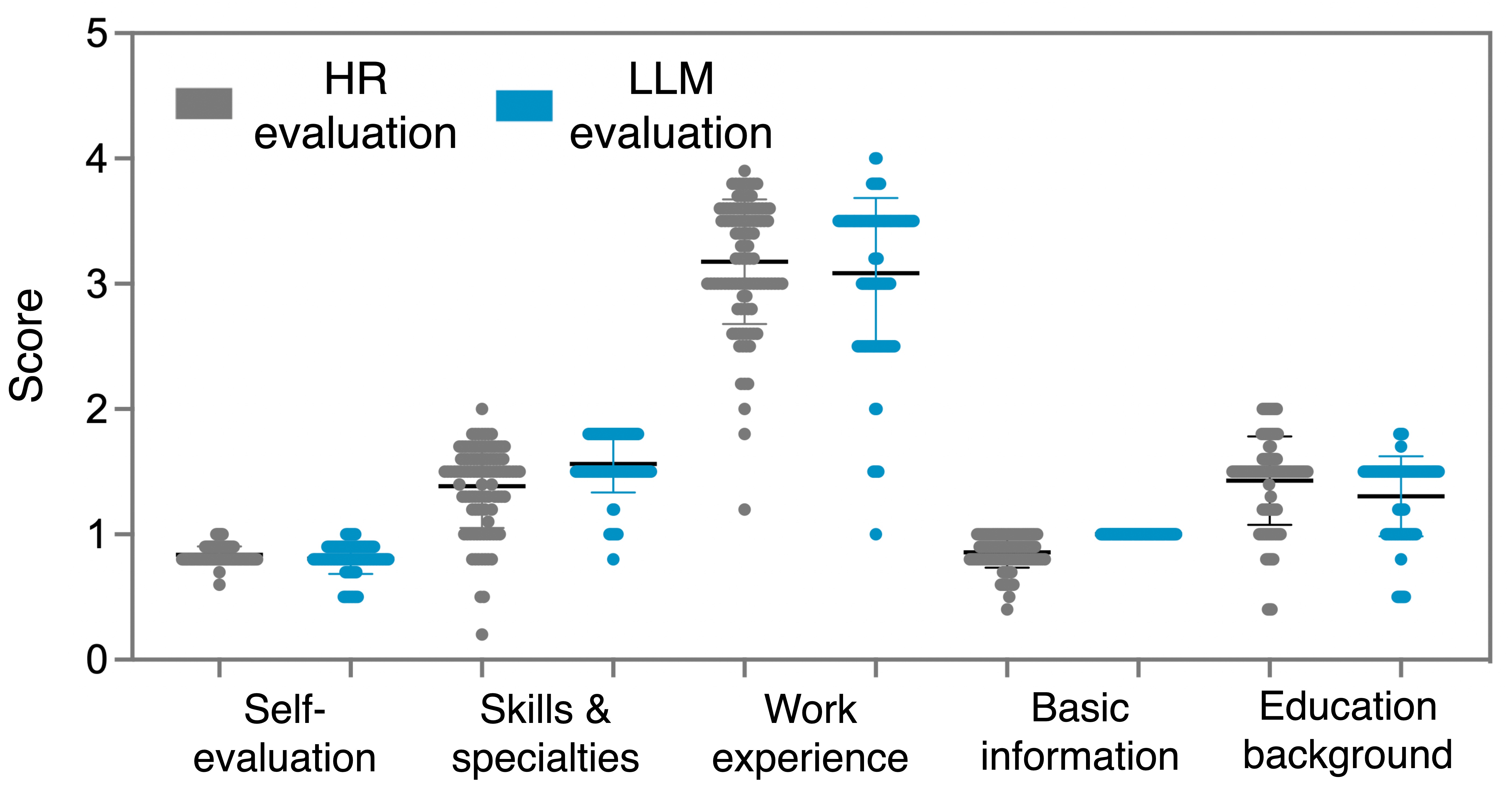}
    \caption{\textbf{Comparison of candidate scores estimated by human evaluators (HR) and RAG-LLM (DeepSeek-V3) across different resume attributes.} The scatter plot visualizes the distribution of scores across five main categories.}
    \label{separate}
    \vspace{-10pt}
\end{figure}

\subsection{Performance of multi-agent RAG-LLMs}
\subsubsection{Comparison with single model approaches}
To evaluate the effectiveness of the proposed multi-agent RAG-LLMs, we first compare their performance against single LLMs across multiple evaluation metrics. Table \ref{table1} presents results using different LLM backbones, including GPT-4o and DeepSeek-V3. The results demonstrate that our proposed RAG-LLM framework achieves satisfactory performance and consistently outperforms single LLMs, confirming its robustness and reliability in AI-driven resume screening. Our evaluation focuses on candidates whose ground truth scores fall within the top and bottom 10\%, 15\%, and 20\% percentiles, enabling a more nuanced analysis of ranking performance under varying selection thresholds. As shown in Table \ref{table1}, our RAG-LLM with DeepSeek-V3 achieves the highest Pearson correlation ($PC_{10}=0.84$, $p$-value $<0.001$), Spearman correlation ($SC_{10}=0.74$, $p$-value $<0.001$), and lowest MAE (0.90), outperforming single LLM baselines. Similar trends persist across the 15\% and 20\% thresholds, highlighting RAG-LLM’s consistent ability to accurately differentiate top-tier candidates from lower-performing ones, ensuring stable and reliable assessments. For borderline candidates, discrepancies between human evaluations and LLM predictions may arise due to subjective judgment, but such variations are expected and fall within a reasonable margin.

\subsubsection{Evaluating the impact of extraction agent}

We further conducted an ablation study to assess the impact of the resume extraction agent, as presented in Table \ref{ablation}. The results show that incorporating structured extraction consistently improves Pearson correlation (PC) and Spearman correlation (SC) across all evaluation thresholds, with DeepSeek-V3 achieving the highest performance when using the extraction module. These findings highlight the importance of high-quality structured resume parsing in enhancing LLM-based candidate evaluations.

\subsubsection{Comparison of AI and human resume screening}

To assess the alignment between human evaluators (HR) and the RAG-LLM model, we analyze the score distributions of both systems. Figure \ref{histogram}(a) presents a scatter plot comparing candidate scores assigned by HR and RAG-LLM (DeepSeek-V3), where the mean scores remain close (i.e., $7.68$ for HR vs. $7.76$ for LLM), indicating high agreement. Figure \ref{histogram}(b) further illustrates this distribution through a histogram, showing that the number of candidates in each score range follows a similar pattern between HR and LLM. These results suggest that RAG-LLM not only achieves strong correlation with human evaluations but also maintains score distribution consistency, reinforcing its reliability for AI-driven hiring applications. Besides, we evaluate the alignment between HR and RAG-LLM assessments across different resume attributes, as shown in Figure \ref{separate}, which compares the score distributions for self-evaluation, skills \& specialties, work experience, basic information, and education background.

\subsection{Qualitative analysis of feedback system}

\begin{figure*}[t]
    \centering
    \includegraphics[width=0.9\linewidth]{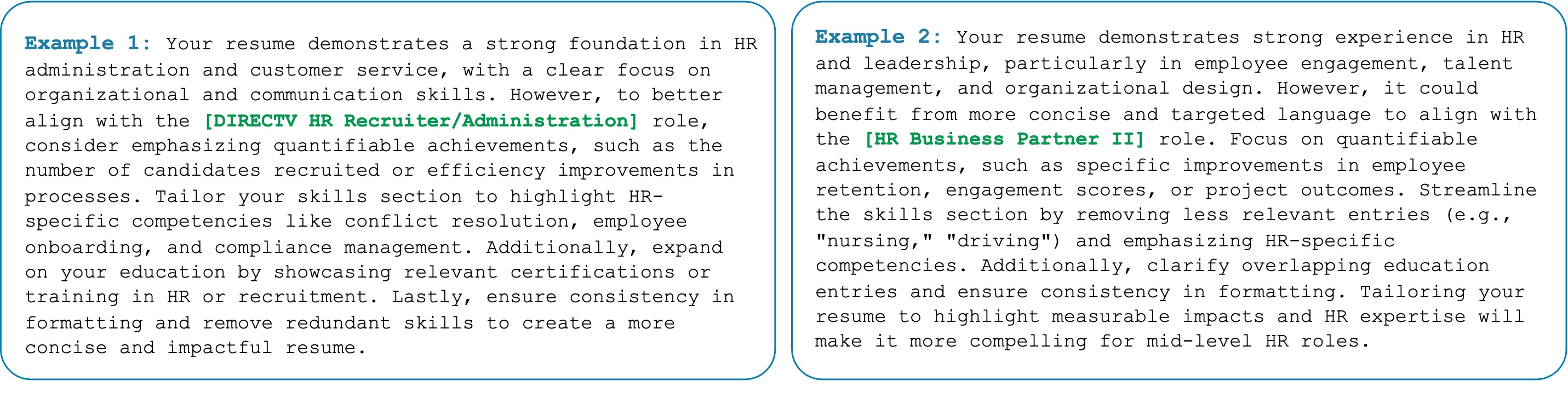}
    \caption{\textbf{Qualitative analysis of resume screening feedback.} Targeted recommendations generated by the summarizer agent after internal discussion among multiple sub-agents (i.e., CEO, CTO, HR agent).}
    \label{feedback}
    \vspace{-10pt}
\end{figure*}

As shown in Figure \ref{feedback} , the summarizer agent consolidates insights from multiple sub-agents, allowing for traceable evaluations. This process reduces recruiter workload by highlighting key strengths and pinpointing missing competencies, eliminating the need for manual resume reviews. Additionally, the system dynamically adapts feedback to different job roles, ensuring that recommendations align with position-specific requirements. To further improve usability, these insights can even be presented in bullet points that summarize strengths and weaknesses, allowing recruiters to efficiently compare multiple candidates.

\section{Discussion}
With the advancement of LLMs and multi-agent systems, AI-driven resume screening has become more effective and reliable than ever. Our study demonstrates that a multi-agent approach offers several advantages over traditional single deep learning models or single LLM-driven screening, particularly in terms of explainability, decision efficiency, and evaluation reliability. One of the most notable findings is that the modular architecture of our system potentially enhances transparency and explainability in AI-driven resume screening. Unlike single model approaches, where recruiters receive only a final score without insight into the reasoning process, our system decomposes resume evaluation into multiple specialized agents. This modularity allows for step-by-step tracking of how each extracted information contributes to the final assessment, improving overall decision accountability. Furthermore, our study highlights an important technical consideration. The extraction agent has a measurable impact on the assessment quality. Specifically, we found that extraction can enhance evaluation by structuring the input data more effectively. However, its effectiveness depends on the model’s reasoning ability, particularly in determining what information should be extracted. Models with stronger reasoning capabilities and a larger number of parameters tend to perform better in this step. This suggests that model selection is crucial, especially in systems where the quality of extraction directly influences the reliability of post-extraction evaluation.

\section{Future Work}

In future work, the integration of multimodal data in LLM-driven hiring has great potential. Current LLM-driven hiring systems mainly focus on text-based resume evaluation, limiting the assessment of soft skills and communication abilities. Incorporating LLM-driven video interview analysis alongside textual evaluation could further provide a more comprehensive assessment of candidate suitability. The system may also generate suitable aptitude and attitude tests to validate a candidate’s actual capabilities (i.e., to verify whether the individual can truly perform the skills or tasks they claim to possess). Apart from this, bias in AI-driven hiring remains a critical concern due to imbalanced training data (e.g., certain demographic groups are underrepresented in the dataset). RAG presents a potential solution by enabling the dynamic retrieval of diverse and up-to-date hiring criteria, reducing reliance on static, historically biased datasets. Future work should explore bias-aware retrieval mechanisms and ranking strategies to enhance the equity and transparency of automated evaluations. Last, privacy considerations in AI-driven hiring remain a critical area for future research, especially as LLMs inference APIs become integral to downstream applications. While external APIs enhance model effectiveness, they also introduce risks of exposing sensitive data to third-party providers. Companies with sufficient computational resources may opt for local LLM deployment to reduce these risks. The future of AI-driven hiring will likely focus on privacy-preserving architectures that enable API-based inference while ensuring compliance with data protection regulations (e.g., GDPR, CCPA). End-to-end encrypted inference techniques, enabling LLMs to compute without directly accessing sensitive data, along with Model Context Protocol (MCP) for structured data flow and context management, are emerging as key research directions. Their integration is expected to play a crucial role in developing secure, scalable, and legally compliant AI-driven hiring systems in the future.

\section{Conclusion}

In this work, we proposed a multi-agent framework for resume screening using RAG-LLMs. The framework is designed with four core agents that work together to extract key resume information, evaluate candidates based on predefined scoring criteria, generate a concise evaluation summary, and format the output in a structured manner. By leveraging RAG, the system can assess resumes against company-specific scoring criteria in a context-aware and tailored manner without requiring model retraining or fine-tuning. To evaluate the effectiveness of our approach, we tested the model using online resume datasets and compared its performance against HR evaluations. The results demonstrated that our proposed framework achieved comparable performance to human evaluators, highlighting the potential of LLMs as an alternative solution for automated and scalable AI hiring.

{
    \small
    \bibliographystyle{unsrtnat}

    \bibliography{main}
}


\end{document}